\documentclass[runningheads]{llncs}

\usepackage{adjustbox}
\usepackage{algorithm}
\usepackage{algpseudocode}
\usepackage{amsmath}
\usepackage{amssymb}
\usepackage{biblatex}
\usepackage{bm}
\usepackage{enumitem}
\usepackage{graphicx}
\usepackage[misc]{ifsym}
\usepackage[frozencache=true,cachedir=minted-cache]{minted}
\usepackage{minted}
\usepackage{multicol}
\usepackage{multirow}
\usepackage{stmaryrd}
\usepackage{subfigure}
\usepackage{pifont}
\usepackage{xr-hyper}

\newcommand{\ie}{\textit{i}.\textit{e}., }

\newcommand{\xmark}{\ding{55}}

\begin{document}

\title{Bridging Few-Shot Learning and Adaptation: \\ New Challenges of Support-Query Shift}

\titlerunning{Bridging Few-Shot Learning and Adaptation}

\toctitle{Bridging Few-Shot Learning and Adaptation:  New Challenges of Support-Query Shift}

\author{Etienne Bennequin$^\star$\inst{1, 2}(\Letter)  \and 
        Victor Bouvier\thanks{Equal contribution}\inst{1, 3}(\Letter) \and
        Myriam Tami\inst{1} \and 
        Antoine Toubhans \inst{2} \and
        C\'eline Hudelot\inst{1}}
        
\tocauthor{Etienne~Bennequin, 
            Victor~Bouvier, 
            Myriam~Tami, 
            Antoine~Toubhans, 
            C\'eline~Hudelot}

\institute{
Universit\'e Paris-Saclay, CentraleSup\'elec, Math\'ematiques et Informatique pour la Complexit\'e et les Syst\`emes, 91190, Gif-sur-Yvette, France \email{firstname.name@centralesupelec.fr} \and
Sicara, 48 boulevard des Batignolles, 75017, Paris, France, \email{etienneb@sicara.com} \and 
Sidetrade, 114 Rue Gallieni, 92100, Boulogne-Billancourt, France, \email{vbouvier@sidetrade.com}}

% \institute{
% Universit\'e Paris-Saclay, CentraleSup\'elec, France, \email{firstname.name@centralesupelec.fr} \and
% Sicara, France, \email{etienneb@sicara.com} \and 
% Sidetrade, France, \email{vbouvier@sidetrade.com}}

\authorrunning{E. Bennequin et al.}
%\author{Anonymous submission 234}
%
%\authorrunning{Anonymous submission 234}
%\institute{}
%
\maketitle              % typeset the header of the contribution
\setcounter{footnote}{0}

\begin{abstract}
\textit{Few-Shot Learning} (FSL) algorithms have made substantial progress in learning novel concepts with just a handful of labelled data. To classify \textit{query} instances  from novel classes encountered at test-time, they only require a \textit{support set} composed of a few labelled samples. FSL benchmarks commonly assume that those queries come from the same distribution as instances in the support set. However, in a realistic setting, data distribution is plausibly subject to change, a situation referred to as \textit{Distribution Shift} (DS). The present work addresses the new and challenging problem of \textit{\textbf{F}ew-Shot Learning under \textbf{S}upport/\textbf{Q}uery \textbf{S}hift} (\textbf{FSQS}) \ie when support and query instances are sampled from related but different distributions. Our contributions are the following. First, we release a testbed for FSQS, including datasets, relevant baselines and a protocol for a rigorous and reproducible evaluation. Second, we observe that well-established FSL algorithms unsurprisingly suffer from a considerable drop in accuracy when facing FSQS, stressing the significance of our study. Finally, we show that transductive algorithms can limit the inopportune effect of DS. In particular, we study both the role of Batch-Normalization and Optimal Transport (OT) in aligning distributions, bridging \textit{Unsupervised Domain Adaptation} with FSL. This results in a new method that efficiently combines OT with the celebrated \textit{Prototypical Networks}. We bring compelling experiments demonstrating the advantage of our method. Our work opens an exciting line of research by providing a testbed and strong baselines. Our code is available at \url{https://github.com/ebennequin/meta-domain-shift}.
\end{abstract}

\section{Introduction}

In the last few years, we have witnessed outstanding progress in supervised deep learning \parencite{he2016deep}. As the abundance of labelled data during training is rarely encountered in practice, ground-breaking works in \textit{Few-Shot Learning} (FSL) have emerged \parencite{Vinyals16,Snell17,Finn17}, particularly for image classification. This paradigm relies on a straightforward setting. At test-time, given a set of not seen during training and \textit{few} (typically 1 to 5) labelled examples for each of those classes, the task is to classify query samples among them. We usually call the set of labelled samples the \textit{support set}, and the set of query samples the \textit{query set}. Well-adopted FSL benchmarks \parencite{Vinyals16,ren2018meta,triantafillou2019meta} commonly sample the support and query sets from the same distribution.  We stress that this assumption does not hold in most use cases. When deployed in the real-world, we expect an algorithm to infer on data that may shift,  resulting in an acquisition system that deteriorates, lighting conditions that vary, or real world objects evolving \parencite{amodei2016concrete}.

\begin{figure}[t]
    \centering
    \subfigure[Standard FSL]{
    \label{fig:SQ_no_shift}
    \includegraphics[width=.475\textwidth]{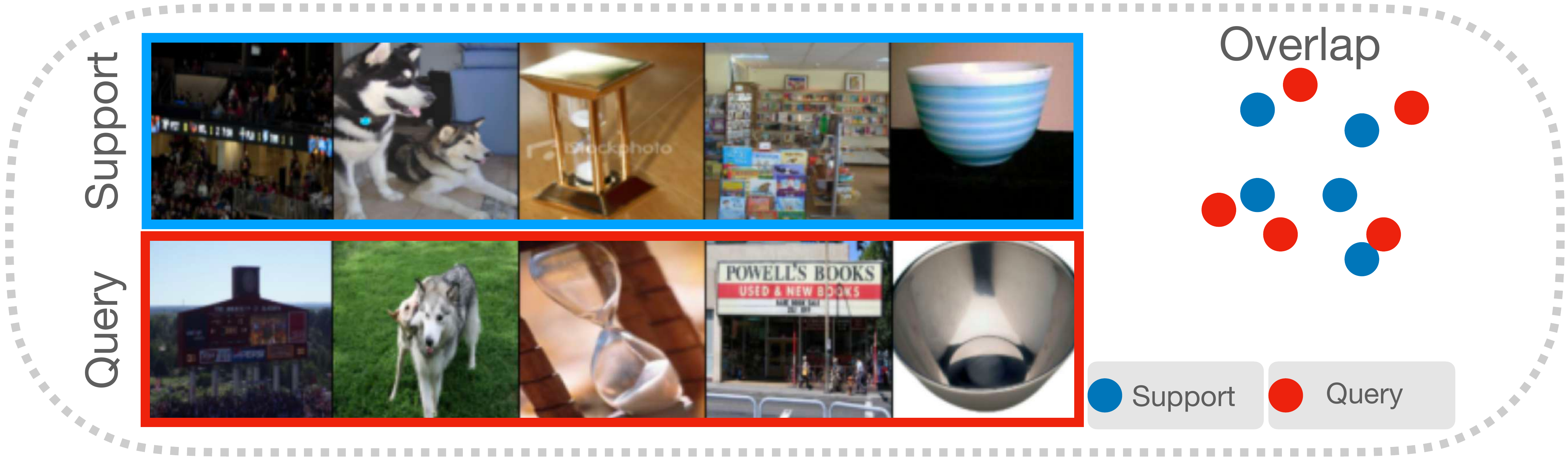}
    }
    \subfigure[FSL under Support / Query Shift]{
    \label{fig:SQ_shift}
    \includegraphics[width=.475\textwidth]{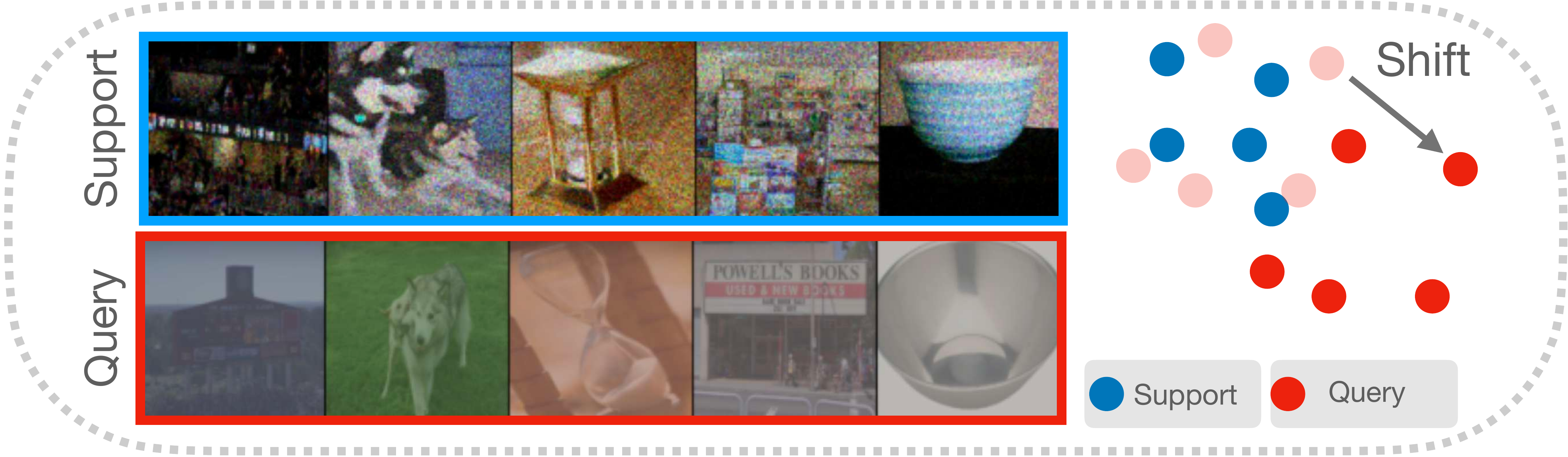}
    } 
    \caption{Illustration of the FSQS problem with a 5-way 1-shot classification task sampled from the miniImageNet dataset \parencite{Vinyals16}. In (a), a standard FSL setting  where support and query sets are sampled from the same distribution. In (b), the same task but with shot-noise and contrast perturbations from \parencite{hendrycks2018benchmarking} applied on support and query sets (respectively) that results in a support-query shift.  In the latter case, a similarity measure based on the Euclidean metric \parencite{Snell17} may become inadequate.}
    \label{fig:support_query_shift}
\end{figure}

The situation of \textit{Distribution Shift} (DS) \ie when training and testing distributions differ, is ubiquitous and has dramatic effects on deep models \parencite{hendrycks2018benchmarking}, motivating works in \textit{Unsupervised Domain Adaptation} \parencite{pan2009survey}, \textit{Domain Generalization} \parencite{gulrajani2021in} or \textit{Test-Time Adaptation} \parencite{wang2020fully}. However, the state of the art brings insufficient knowledge on few-shot learners' behaviours when facing distribution shift. Some pioneering works demonstrate that advanced FSL algorithms do not handle cross-domain generalization better than more naive approaches \parencite{Chen19}. Despite its great practical interest, FSL under distribution shift between the support and query sets %(\ie when they are sampled from related but different distributions) 
is an under-investigated problem and attracts a very recent attention \parencite{du2021metanorm}. We refer to it as \textit{\textbf{F}ew-Shot Learning under \textbf{S}upport/\textbf{Q}uery \textbf{S}hift} (\textbf{FSQS}) and provide an illustration in Figure \ref{fig:support_query_shift}. It reflects a more realistic situation where the algorithm is fed with a support set at the time of deployment and infers continuously on data subject to shift. The first solution is to re-acquire a support set that follows the data's evolution. Nevertheless, it implies human intervention to select and annotate data to update an already deployed model, reacting to a potential drop in performances. The second solution consists in designing an algorithm that is robust to the distribution shift encountered during inference. This is the subject of the present work. Our contributions are:
\begin{enumerate}
    \item \texttt{FewShiftBed}: a testbed for FSQS available at \url{https://github.com/ebennequin/meta-domain-shift}. The testbed includes 3 challenging benchmarks along with a protocol for fair and rigorous comparison across methods as well as an implementation of relevant baselines, and an interface to facilitate the implementation of new methods.

    \item We conduct extensive experimentation of a representative set of few-shot algorithms. We empirically show that \textit{Transductive} Batch-Normalization \parencite{bronskill2020tasknorm} mitigates an important part of the inopportune effect of FSQS.
    \item We bridge \textit{Unsupervised Domain Adaptation} (UDA) with FSL to address FSQS. We introduce \textit{Transported Prototypes}, an efficient transductive algorithm that couples \textit{Optimal Transport} (OT) \parencite{peyre2019computational} with the celebrated \textit{Prototypical Networks} \parencite{Snell17}. The use of OT follows a long-standing history in UDA for aligning representations between distributions \parencite{ben2007analysis,ganin2015unsupervised}. Our experiments demonstrate that OT shows a remarkable ability to perform this alignment even with only a few samples to compare distributions and provide a simple but strong baseline.
\end{enumerate}
In Section~\ref{section:statement_SQS} we provide a formal statement of FSQS, and we position this new problem among existing learning paradigms. In Section \ref{section:few_shift_bed}, we present \texttt{FewShiftBed}. We detail the datasets, the chosen baselines, and a protocol that guarantees a rigorous and reproducible evaluation. In Section \ref{section:OT_prototypes}, we present a method that couples Optimal Transport with Prototypical Networks \parencite{Snell17}. Finally, in Section \ref{section:experiments}, we conduct an extensive evaluation of baselines and our proposed method using the testbed.

\section{The Support-Query Shift problem}

\label{section:statement_SQS}
\subsection{Statement}

\paragraph{Notations.} We consider an input space $\mathcal X$, a representation space $\mathcal Z \subset \mathbb  R^d$ ($d>0$) and a set of classes $\mathsf C$. A representation is a learnable function from $\mathcal X$ to $\mathcal Z$ and is noted $\varphi(\cdot ; \theta)$ with $\theta \in \Theta$ for $\Theta$ a set of parameters. A dataset is a set $\Delta (\mathsf C, \mathsf D)$ defined by a set of classes $\mathsf C$ and a set of domains $\mathsf D$ \ie a domain $\mathcal D \in \mathsf D$ is a set of IID realizations from a distribution noted $p_{\mathcal D}$. For two domains $\mathcal D,\mathcal D' \in \mathsf D$, the distribution shift is characterized by $p_{\mathcal D} \neq p_{\mathcal D'}$. For instance, if the data consists of images of letters handwritten by several users, $\mathcal D$ can consist of samples from a specific user. Referring to the well known UDA  terminology of source / target \parencite{pan2009survey}, we define a couple of source-target domains as a couple $(\mathcal D_s, \mathcal D_t)$ with $p_{\mathcal D_s} \neq p_{\mathcal D_t}$, thus presenting a distribution shift. Additionally, given $\mathcal C \subset\mathsf C$ and $\mathcal D \in \mathsf D$,  the restriction of a domain $\mathcal D$ to images with a label that belongs to $\mathcal C$ is noted $\mathcal D^{\mathcal C}$.

\begin{figure}[t]
    \centering
    \includegraphics[scale=0.17]{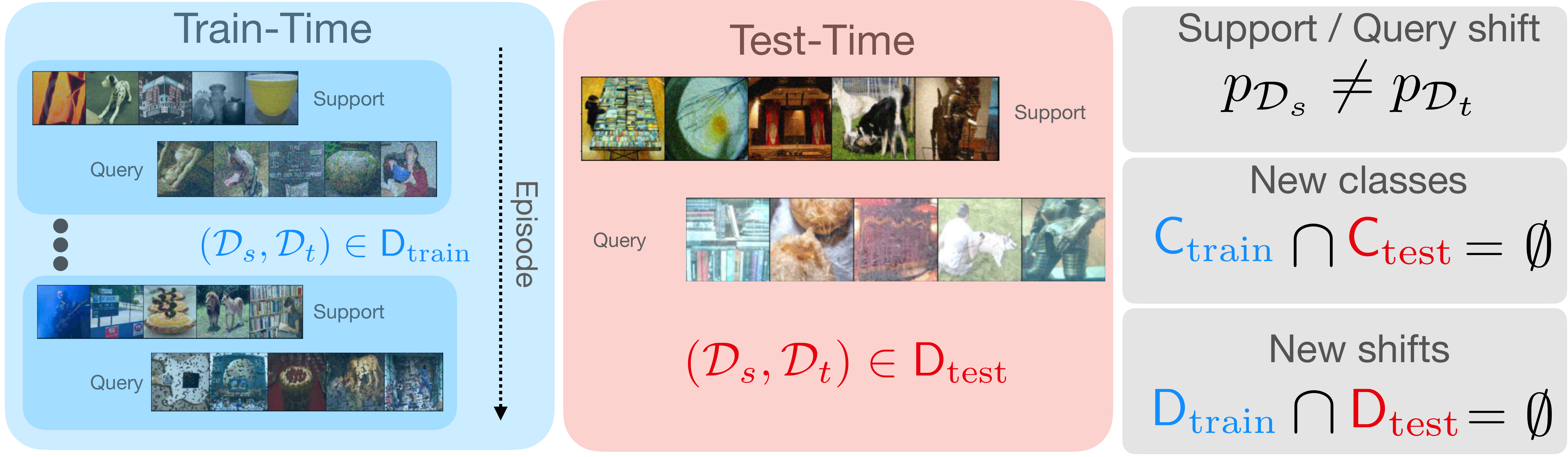}
    \caption{During meta-learning (Train-Time), each episode contains a support and a query set sampled from different distributions (for instance, illustrated by noise and contrasts as in Figure \ref{fig:SQ_shift}) from a set of \textit{training domains} ($\mathsf D_{\mathrm{train}}$), reflecting a situation that may potentially occurs at test-time. When deployed, the FSL algorithm using a trained backbone is fed with a support set sampled from new classes. As the algorithm is subject to infer continuously on data subject to shift (Test-Time), we evaluate the algorithm on data with an unknown shift ($\mathsf D_{\mathrm{test}}$). Importantly, both classes (${\mathsf C}_{\mathrm{train}} \cap {\mathsf C}_{\mathrm{test}} = \emptyset$) and shifts ($\mathsf D_{\mathrm{train}} \cap \mathsf D_{\mathrm{test}} = \emptyset$)  are not seen during training, making the FSQS a challenging problem of generalization.}
    \label{fig:protocol_fsq}
\end{figure}

\paragraph{Dataset splits.} We build a split of $\Delta(\mathsf C, \mathsf D)$, by splitting $\mathsf D$ (respectively $\mathsf C$) into $\mathsf D_{\text{train}}$ and $\mathsf D_{\text{test}}$ (respectively $\mathsf C_{\text{train}}$ and $\mathsf C_{\text{test}}$) such that $\mathsf D_{\text{train}} \cap \mathsf D_{\text{test}} = \emptyset$ and $\mathsf D_{\text{train}} \cup \mathsf D_{\text{test}} = \mathsf D$ (respectively $\mathsf C_{\text{train}} \cap \mathsf C_{\text{test}} = \emptyset$ and $\mathsf C_{\text{train}} \cup \mathsf C_{\text{test}} = \mathsf C$). This gives us a train/test split with the datasets $\Delta_{\text{train}} = \Delta(\mathsf C_{\text{train}}, \mathsf D_{\text{train}})$ and $\Delta_{\text{test}} = \Delta(\mathsf C_{\text{test}}, \mathsf D_{\text{test}})$. By extension, we build a validation set following the same protocol.

\paragraph{Few-Shot Learning under Support-Query Shift (FSQS).} Given:
\begin{itemize}
    \item $\mathsf D' \in \{ \mathsf D_{\mathrm{train}}, \mathsf D_{\mathrm{test}}\}$ and $\mathsf C' \in \{\mathsf C_{\mathrm{train}}, \mathsf C_{\mathrm{test}}\}$,
     \item a couple of source-target domains $(\mathcal D_s, \mathcal D_t)$ from $\mathsf D'$,
    \item a set of classes $\mathcal C \subset \mathsf C'$;
    \item a small labelled support set $\mathcal{S} = {(x_i, y_i)}_{i=1, \dots, |\mathcal{S}|}$ (named \textit{source support set}) such that for all $i$, $y_i \in \mathcal C$ and $x_i \in \mathcal D_s$ \ie $\mathcal{S} \subset \mathcal D_s^{\mathcal C}$;
    \item an unlabelled query set $\mathcal{Q} = {(x_i)}_{i=1, \dots, |\mathcal{Q}|}$ (named \textit{target query set}) such that for all $i$, $y_i \in \mathcal C$ and $x_i \in \mathcal D_t$ \ie $\mathcal{Q} \subset \mathcal D_t^{\mathcal C}$.
\end{itemize}

The task is to predict the labels of query set instances in $\mathcal C$. When $|\mathcal{C}| = n$ and the support set contains $k$ labelled instances for each class, this is called an $n$-way $k$-shot FSQS classification task. %However, the methods studied in this paper are not limited to the case where the support set is perfectly balanced. This later setting suggests that the support set is perfectly balanced which is not a requirement of the methods studied in this paper (or a requirement of our method 
Note that this paradigm provides an additional challenge compared to classical Few-shot classification tasks, since at test time, the model is expected to generalize to both new classes and new domains while support set and query set are sampled from different distributions.
% As presented above, in order to evaluate the model's capacity to generalize to both new classes and new domains, we split the dataset into train, validation and test sets (respectively named $\Delta_{\mathrm{train}}$, $\Delta_{\mathrm{val}}$ and $\Delta_{\mathrm{test}}$) controlling that there is no overlap of both classes and domains of these sets. Model's parameters are trained on $\Delta_{\mathrm{train}}$ and selected on  $\Delta_{\mathrm{val}}$. Finally, the model is tested on $\Delta_{\text{test}}$. 
This paradigm is illustrated in Figure \ref{fig:protocol_fsq}.

\paragraph{Episodic training.}  We build an episode by sampling some classes $\mathcal{C} \subset \mathsf C_{\text{train}}$, and a source and target domain $\mathcal D_s, \mathcal D_t$ from $\mathsf D_{\text{train}}$. We build a support set $\mathcal{S} = {(x_i, y_i)}_{i=1 \dots |\mathcal{S}|}$ of instances from source domain $\mathcal D_s^{\mathcal C}$, and a query set $\mathcal{Q} = {(x_i, y_i)}_{i=|\mathcal{S}|+1 , \dots , |\mathcal{S}|+|\mathcal{Q}|}$ of instances from target domain $\mathcal D_t^{\mathcal C}$, such that $\forall i \in \left[ 1, |\mathcal{S}|+|\mathcal{Q}| \right]$, $y_i \in \mathcal{C}$. Using the labelled examples from $\mathcal{S}$ and unlabelled instances from $\mathcal{Q}$, the model is expected to predict the labels of $\mathcal{Q}$. The parameters of the model are then trained using a cross-entropy loss between the predicted labels and ground truth labels of the query set.

%\paragraph{Episodic training.}  At episode $\tau$, we sample some classes $\mathcal{C}_\tau \subset \mathcal{C_{\text{train}}}$, and a source and target domain $\mathbf{d}_{s,\tau},\mathbf{d}_{t,\tau}$ from $\Gamma_{\text{train}}$. We build a support set $\mathcal{S}_\tau = {(x_i, y_i)}_{i=1 \dots |\mathcal{S}_\tau|}$ of instances from source domain $\mathbf{d}_{s,\tau}^{\mathcal C_{\tau}}$, and a query set $\mathcal{Q}_\tau = {(x_i, y_i)}_{i=|\mathcal{S}_\tau|+1 \dots |\mathcal{S}_\tau|+|\mathcal{Q}_\tau|}$ of instances from target domain $\mathbf{d}_{t, \tau}^{\mathcal C_{\tau}}$, such that $\forall i \in \left[ 1, |\mathcal{S}_\tau|+|\mathcal{Q}_\tau| \right], y_i \in \mathcal{C}_\tau$. Using the labelled examples from $\mathcal{S}_\tau$ and unlabelled instances from $\mathcal{Q}_\tau$, the model is expected to predict the labels of $\mathcal{Q}_\tau$. The parameters of the model are then trained using a classification loss between the predicted labels and ground truth labels of $\mathcal{Q}_\tau$. We provide more training details in the following.

\subsection{Positioning and Related Works}
\label{section:positioning}

To highlight FSQS's novelty, our discussion revolves around the problem of inferring on a given \textit{Query Set} provided with the knowledge of a \textit{Support Set}. We refer to this class of problems as \textit{SQ problems}. Intrinsically, FSL falls into the category of SQ problems. Interestingly, \textit{Unsupervised Domain Adaptation} \parencite{pan2009survey} (UDA), defined as labelling a dataset sampled from a target domain based on labelled data sampled from a source domain, is also a SQ problem. Indeed, in this case, the source domain plays the role of support, while the target domain plays the query's role. Notably, an essential line of study in UDA leverages the target data distribution for aligning source and target domains, reflecting the importance of transduction in a context of adaptation \parencite{ben2007analysis,ganin2015unsupervised} \ie performing prediction by considering all target samples together. Transductive algorithms also have a special place in FSL \parencite{dhillon2019baseline,liu2018learning,ren2018meta} and show that leveraging a query set as a whole brings a significant boost in performances. Nevertheless, UDA and FSL exhibit fundamental differences. UDA addresses the problem of distribution shift using important source data and target data (typically thousands of instances) to align distributions. In contrast, FSL focuses on the difficulty of learning from few samples. To this purpose, we frame UDA as both SQ problem with \textit{large} transductivity and Support / Query Shift, while Few-Shot Learning is a SQ problem, eventually with \textit{small} transductivity for transductive FSL. Thus, FSQS combines both challenges: distribution shift and small transductivity. This new perspective allows us to establish fruitful connections with related learning paradigms, presented in Table \ref{table:paradigm}, that we review in the following. A thorough review is available in Appendix A\footnote{\url{https://arxiv.org/abs/2105.11804}}.

\begin{table}[t]
  \begin{center}
  \begin{adjustbox}{max width=\textwidth}
  \begin{tabular}{|cl|cc|cc|cc|c|c|c|}
    \hline 
      &  \multirow{2}{*}{\textbf{SQ problems}} &  \multicolumn{4}{|c|}{\textbf{Train-Time}} & \multicolumn{5}{|c|}{\textbf{Test-Time}}   \\
    & & \multicolumn{2}{|c|}{Support}  & \multicolumn{2}{|c|}{Query} & \multicolumn{2}{|c|}{Support} & \multicolumn{1}{|c|}{Query} &  \footnotesize{New} & \footnotesize{New}  \\
    & & \scriptsize{Size} & \scriptsize{Labels} & \scriptsize{Size} & \scriptsize{Labels} & \scriptsize{Size} & \scriptsize{Labels} & \scriptsize{Transductivity} & \footnotesize{classes} & \footnotesize{domains}   \\
    \hline \hline 
     ~ \multirow{3}{*}{\rotatebox[origin=c]{90}{\scriptsize{\textbf{No SQS}}}}~ & \footnotesize{FSL \parencite{Snell17,Finn17}} & \scriptsize{Few} & \checkmark &  \scriptsize{Few} & \checkmark  &  \scriptsize{Few} & \checkmark &  \scriptsize{Point-wise} &  \checkmark & \xmark    \\
    & \footnotesize{TransFSL \parencite{ren2018meta,liu2018learning}}  & \scriptsize{Few} & \checkmark &  \scriptsize{Few} & \checkmark  &  \scriptsize{Few} & \checkmark &  \scriptsize{Small  } &  \checkmark & \xmark   \\
    & \footnotesize{CDFSL \parencite{Chen19}} & \scriptsize{Few} & \checkmark &  \scriptsize{Few} & \checkmark  &  \scriptsize{Few} & \checkmark &  \scriptsize{Point-wise} &  \checkmark & \checkmark \\
    \hline
    ~ \multirow{5}{*}{\rotatebox[origin=c]{90}{\textbf{SQS}}}~ & \footnotesize{UDA \parencite{quionero2009dataset,pan2009survey}} & & & & &    \scriptsize{Large} & \checkmark &  \scriptsize{Large} &  &   \\
    
    %& \footnotesize{Source data free UDA \parencite{liang2020we,yehsofa}} & \scriptsize{Large} & \checkmark & & &  &   &  \scriptsize{Large  } &  &    \\
    
    & \footnotesize{TTA \parencite{sun2020test,schneider2020improving,wang2020fully}} & \scriptsize{Large} & \checkmark &  &  &   &  & \scriptsize{Small  } & & \checkmark  \\
    & \footnotesize{ARM \parencite{zhang2020adaptive}} & \scriptsize{Large} & \checkmark &  \scriptsize{Few} & \checkmark  &   &  & \scriptsize{Small  } &  & \checkmark  \\
    & \footnotesize{Ind FSQS}  & \scriptsize{Few} & \checkmark &  \scriptsize{Few} & \checkmark &   \scriptsize{Few} & \checkmark  & \scriptsize{Point-wise } & \checkmark & \checkmark   \\
    & \footnotesize{Trans FSQS}  & \scriptsize{Few} & \checkmark &  \scriptsize{Few} & \checkmark &   \scriptsize{Few} & \checkmark  & \scriptsize{Small  } & \checkmark & \checkmark   \\
    \hline 
  \end{tabular}
  \end{adjustbox}
  \end{center}    
  \caption{An overview of SQ problems. We divide SQ problems into two categories, presence or not of \textbf{S}upport-\textbf{Q}uery shift; \textbf{No SQS} \textit{vs} \textbf{SQS}. We consider three classes of transductivity: point-wise transductivity that is equivalent to inductive inference, small transductivity when inference is performed at batch level (typically in \parencite{wang2020fully,zhang2020adaptive}), and large transductivity when inference is performed at dataset level (typically in UDA). New classes (resp. new domains) describe if the model is evaluated at test-time on novel classes (resp. novel domains). Note that we frame UDA as a fully test-time algorithm. Notably, Cross-Domain FSL (CDFSL) \parencite{Chen19} assumes that the support set and query set are drawn from the same distribution, thus No SQS.}
  \label{table:paradigm}  
\end{table}

\paragraph{Adaptation.} Unsupervised Domain Adaptation (UDA) requires a whole target dataset for inference, limiting its applications. Recent pioneering works, referred to as Test-Time Adaptation (TTA), adapt at test-time a model provided with a batch of samples from the target distribution. The proposed methodologies are test-time training by self-supervision \parencite{sun2020test}, updating  batch-normalization statistics \parencite{schneider2020improving} or parameters  \parencite{wang2020fully}, or meta-learning to condition predictions on the whole batch of test samples for an \textit{Adaptative Risk Minimization} (ARM) \parencite{zhang2020adaptive}. Inspired from the principle of invariant representations \parencite{ben2007analysis,ganin2015unsupervised}, the seminal work \parencite{courty2016optimal} brings \textit{Optimal Transport} (OT)~\parencite{peyre2019computational} as an efficient framework for aligning data distributions. OT has been recently applied in a context of transductive FSL \parencite{hu2020leveraging} and our proposal (TP) is to provide a simple and strong baseline following the principle of OT as it is applied in UDA.  In this work, following \parencite{bronskill2020tasknorm}, we also study the role of Batch-Normalization for SQS, that points out the role of transductivity. Our conviction was that the batch-normalization is the first lever for aligning distributions \parencite{schneider2020improving,wang2020fully}.

\paragraph{Few-Shot Classification.} We usually frame Few-Shot Classification methods \parencite{Chen19} as either metric-based methods \parencite{Vinyals16,Snell17}, or optimization-based methods that learn to fine-tune by adapting with few gradient steps \parencite{Finn17}. A promising line of study leverages \textit{transductivity}  (using the query set as unlabelled data while inductive methods predict individually on each query sample). Transductive Propagation Network \parencite{liu2018learning} meta-learns label propagation from the support to query set concurrently with the feature extractor. Transductive Fine-Tuning \parencite{dhillon2019baseline} minimizes the prediction entropy of all query instances during fine-tuning. Evaluating cross-domain generalization of FSL (FSCD), \ie a distributional shift between meta-training and meta-testing, attracts the attention of a few recent works \parencite{Chen19}. Zhao \textit{et al.} propose a Domain-Adversarial Prototypical Network \parencite{zhao2020domain} in order to both align source and target domains in the feature space while maintaining discriminativeness between classes. Sahoo \textit{et al.} combine Prototypical Networks with adversarial domain adaptation at the task level \parencite{sahoo2019meta}. Notably, Cross-Domain Few-Shot Learning \parencite{Chen19} (CDFSL) addresses the distributional shift between meta-training and meta-testing assuming that the support set and query set are drawn from the same distribution, not making it a SQ problem with support-query shift. Concerning the novelty of FSQS, we acknowledge the very recent contribution of Du \textit{et al.} \parencite{du2021metanorm} which studies the role of learnable normalization for domain generalization, in particular when support and query sets are sampled from different domains. Note that our statement is more ambitious: we evaluate algorithms on both source and target domains that were unseen during training, while in their setting the source domain has already been seen during training. 

\paragraph{Benchmarks in Machine Learning} Releasing benchmark has always been an important factor for progress in the \textit{Machine Learning} field, the most outstanding example being ImageNet \parencite{deng2009imagenet} for the Computer Vision community. Recently, \texttt{DomainBed}  \parencite{gulrajani2021in} aims to settle Domain Generalization research into a more rigorous process, where  \texttt{FewShiftBed} takes inspiration from it. \texttt{Meta-Dataset} \parencite{triantafillou2019meta} is an other example, this time specific to FSL.

\section{\texttt{FewShiftBed}: A \texttt{Pytorch} testbed for FSQS}
\label{section:few_shift_bed}
\subsection{Datasets}
We designed three new image classification datasets adapted to the FSQS problem. These datasets have two specificities.

\begin{enumerate}
    \item They are dividable into groups of images, assuming that each group corresponds to a distinct domain. A key challenge is that each group must contain enough images with a sufficient variety of class labels, so that it is possible to sample FSQS episodes.
    \item They are delivered with a train/val/test split ($\Delta_{\text{train}}, \Delta_{\text{val}}, \Delta_{\text{test}}$), along both the class and the domain axis. This split is performed following the principles detailed in Section \ref{section:statement_SQS}. Therefore, these datasets provide true few-shot tasks at test time, in the sense that the model will not have seen any instances of test classes and domains during training. Note that since we split along two axes, some data may be discarded (for instance images from a domain in $\mathsf D_{\text{train}}$ with a label in $\mathsf C_{\text{test}}$). Therefore it is crucial to find a split that minimizes this loss of data.
\end{enumerate}

\paragraph{Meta-CIFAR100-Corrupted (MC100-C).} CIFAR-100 \parencite{krizhevsky2009learning} is a dataset of 60k three-channel square images of size $32 \times 32$, evenly distributed in 100 classes. Classes are evenly distributed in 20 superclasses. We use the same method used to build CIFAR-10-C \parencite{hendrycks2018benchmarking}, which makes use of 19 image perturbations, each one being applied with 5 different levels of intensity, to evaluate the robustness of a model to domain shift. We modify their protocol to adapt it to the FSQS problem: (i) we split the classes with respect to the superclass structure, and assign 13 superclasses (65 classes) to the training set, 2 superclasses (10 classes) to the validation set, and 5 superclasses (25 classes) to the testing set; (ii) we also split image perturbations (acting as domains), following the split of \parencite{zhang2020adaptive}. We obtain 2,184k transformed images for training, 114k for validation and 330k for testing. The detailed split is available in the documentation of our code repository.

\paragraph{\textit{mini}ImageNet-Corrupted (mIN-C).} \textit{mini}ImageNet \parencite{Vinyals16} is a popular benchmark for few-shot image classification. It contains 60k images from 100 classes from the ImageNet dataset. 64 classes are assigned to the training set, 16 to the validation set and 20 to the test set. Like MC100-C, we build mIN-C using the image perturbations proposed by \parencite{hendrycks2018benchmarking} to simulate different domains. We use the original split from \parencite{Vinyals16} for classes, and use the same domain split as for MC100-C. Although the original \textit{mini}ImageNet uses $84 \times 84$ images, we use $224 \times 224$ images. This allows us to re-use the perturbation parameters calibrated in \parencite{hendrycks2018benchmarking} for ImageNet. Finally, we discard the 5 most time-consuming perturbations. We obtain a total of 1.2M transformed images for training, 182k for validation and 228k for testing. The detailed split in the documentation of our code repository.

\paragraph{FEMNIST-FewShot (FEMNIST-FS).} EMNIST \parencite{cohen2017emnist} is a dataset of images of handwritten digits and uppercase and lowercase characters. Federated-EMNIST \parencite{caldas2018leaf} is a version of EMNIST where images are sorted by writer (or user). FEMNIST-FS consists in a split of the FEMNIST dataset adapted to few-shot classification. We separate both users and classes between training, validation and test sets. We build each group as the set of images written by one user. The detailed split is available in the code. Note that in FEMNIST, many users provide several instances for each digits, but less than two instance for most letters. Therefore it is hard to find enough samples from a user to build a support set or a query set. As a result, our experiments are limited to classification tasks with only one sample per class in both the support and query sets.

\subsection{Algorithms}

We implement in \texttt{FewShiftBed} two representative methods of the vast literature of FSL, that are commonly considered as strong baselines: Prototypical Networks (\textbf{ProtoNet}) \parencite{Snell17} and Matching Networks  (\textbf{MatchingNet}) \parencite{Vinyals16}. Besides, for transductive FSL, we also implement with Transductive Propagation Network (\textbf{TransPropNet}) \parencite{liu2018learning} and Transductive Fine-Tuning (\textbf{FTNet}) \parencite{dhillon2019baseline}. We also implement our novel algorithm \textit{Transported Prototypes} (\textbf{TP}) which is detailed in Section \ref{section:OT_prototypes}. \texttt{FewShiftBed} is designed for favoring a straightforward implementation of a new algorithm for FSQS. To add a new algorithm, we only need to implement the \textcolor{blue}{\texttt{set\_forward}} method of the class \textcolor{blue}{\texttt{AbstractMetaLearner}}. We provide an example with our implementation of the Prototypical Network \parencite{Snell17} that only requires few line of codes:

\begin{center}
\scriptsize{
    \begin{minted}[firstnumber=last]{python}
    class ProtoNet(AbstractMetaLearner):
        def set_forward(self, support_images, support_labels, query_images):
            z_support, z_query = self.extract_features(support_images, query_images)
            z_proto = self.get_prototypes(z_support, support_labels)
            return  - euclidean_dist(z_query, z_proto)
    \end{minted}
}
\end{center}

\subsection{Protocol}
\label{section:protocol}
To prevent the pitfall of misinterpreting a performance boost, we draw three recommendations to isolate the causes of improvement rigorously.
\begin{itemize}
    \item \textbf{How important is episodic training?} Despite its wide adoption in meta-learning for FSL, in some situation episodic training does not perform better than more naive approaches \parencite{Chen19}. Therefore we recommend to report both the result obtained using episodic training and standard ERM (see the documentation of our code repository).
    \item \textbf{How does the algorithm behave in the absence of Support-Query Shift?} In order to assess that an algorithm designed for distribution shift does not provide degraded performance in an ordinary concept, and to provide a top-performing baseline, we recommend reporting the model's performance when we do not observe, at test-time, a support-query shift. Note that it is equivalent to evaluate the performance in cross-domain generalization, as firstly described in \parencite{Chen19}. 
    \item \textbf{Is the algorithm transductive?} The assumption of transductivity has been responsible of several improvements in FSL \parencite{ren2018meta,bronskill2020tasknorm} while it has been demonstrated in \parencite{bronskill2020tasknorm} that MAML \parencite{Finn17} benefits strongly from the Transductive Batch-Normalization (TBN). Thus, we recommend specifying if the method is transductive and adapting the choice of the batch-normalization accordingly (Conventional Batch Normalization \parencite{ioffe2015batch} and Transductive Batch Normalization for inductive and transductive methods, respectively) since transductive batch normalization brings a significant boost in performance \parencite{bronskill2020tasknorm}.
\end{itemize}

\section{Transported Prototypes: A baseline for FSQS}
\begin{figure}[t]
    \centering
    \includegraphics[scale=0.22]{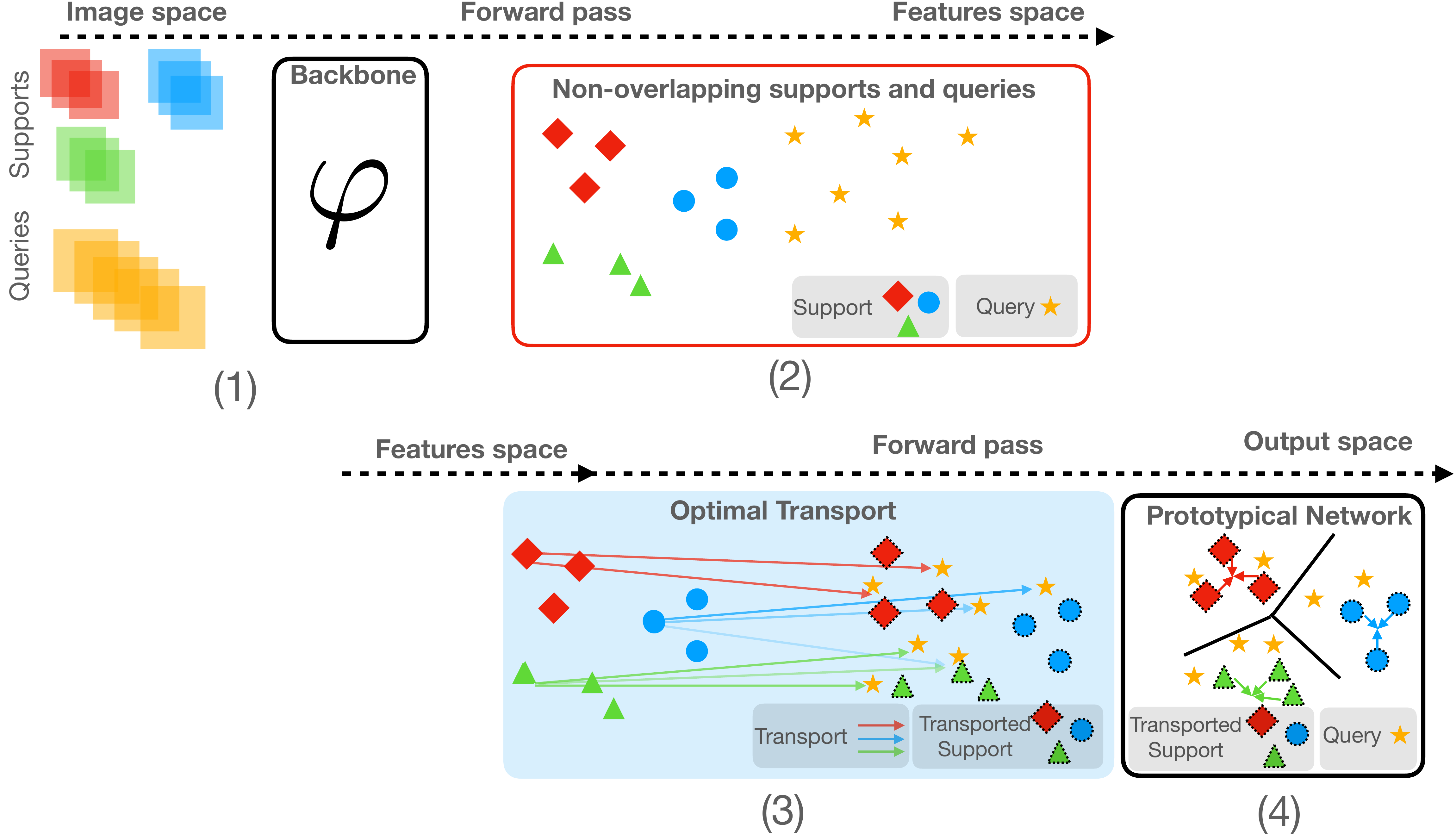}
    \caption{Overview of \textit{Transported Prototypes}. (1) A support set and a query set are fed to a trained backbone that embeds images into a feature space. (2) Due to the shift between distributions, support and query instances are embedded in non-overlapping areas. (3) We compute the Optimal Transport from support instances to query instances to build the transported support set. Note that we represent the transport plan only for one instance per class to preserve clarity in the schema. (4) Provided with the transported support, we apply the Prototypical Network \parencite{Snell17} \ie $L^2$ similarity between transported support and query instances.}
    \label{fig:transported_prototypes}
\end{figure}
\label{section:OT_prototypes}

\subsection{Overall idea}
We present a novel method that brings UDA to FSQS. As aforementioned, FSQS presents new challenges since we no longer assume that we sample the support set and the query set from the same distribution. As a result, it is unlikely that the support set and query sets share the same representation space region (non-overlap). In particular, the $L^2$ distance, adopted in the celebrated Prototypical Network \parencite{Snell17}, may not be relevant for measuring similarity between query and support instances, as presented in Figure \ref{fig:support_query_shift}. To overcome this issue, we develop a two-phase approach that combines Optimal Transport (Transportation Phase) and the celebrated Prototypical Network (Prototype Phase). We give some background about Optimal Transport (OT) in Section \ref{section:background} and the whole procedure is presented in Algorithm \ref{alg:transported_prototypes}.

\subsection{Background}
\label{section:background}
\paragraph{Definition.} We provide some basics about Optimal Transport (OT). A thorough presentation of OT is available at \parencite{peyre2019computational}. Let $p_s$ and $p_t$ be two distributions on $\mathcal X$, we note $\Pi(p_s, p_t)$ the set of joint probability with marginal $p_s$ and $p_t$ \ie $\forall \pi \in \Pi(p_s, p_t), \forall x \in \mathcal X, \pi(\cdot, x) = p_s, \pi(x, \cdot) = p_t$. The \textit{Optimal Transport}, associated to cost $c$, between $p_s$ and $p_t$ is defined as:
\begin{equation}
    W_c(p_s, p_t) := \min_{\pi \in \Pi(p_s, p_t)} \mathbb E_{(x_s,x_t) \sim \pi} \left [ c(x_s,x_t)\right]
    \label{eq:OT_def}
\end{equation}
with $c(\cdot,\cdot)$ any metric. We note $\pi^\star(p_s, p_t)$ the joint distribution that achieves the minimum in equation \ref{eq:OT_def}. It is named the \textit{transportation plan} from $p_s$ to $p_t$. When there is no confusion, we simply note $\pi^\star$. For our applications, we will use as metric the euclidean distance in the representation space obtained from a representation $\varphi(\cdot; \theta)$ \ie $c_\theta(x_s,x_t) := ||\varphi(x_s; \theta)-\varphi(x_t; \theta)||_2$.

\paragraph{Discrete OT.} When $p_s$ and $p_t$ are only accessible through a finite set of samples, respectively $(x_{s,1},..., x_{s,n_s})$ and $(x_{t, 1},...,x_{t, n_t})$ we introduce the empirical distributions $
\hat p_s := \sum_{i=1}^{n_s} w_{s,i} \delta_{x_{s,i}}, ~~ \hat p_t := \sum_{j=1}^{n_t} w_{t,j} \delta_{x_{t,j}}$, where $w_{s,i}$ ($w_{t,j}$) is the mass probability put in sample $x_{s,i}$ ($x_{t,j}$) \ie $\sum_{i=1}^{n_s} w_{s,i}=1$ ($\sum_{j=1}^{n_t} w_{t,j}=1$) and $\delta_x$ is the Dirac distribution in $x$. The discrete version of the OT is derived by introducing the set of couplings $
    \mathbf \Pi(p_s,p_t) := \left \{ \bm \pi \in \mathbb R^{n_s\times n_t}, \bm \pi \mathbf 1_{n_s} = \mathbf p_s, \bm \pi^\top \mathbf 1_{n_t} = \mathbf p_t \right \}
$ where $\mathbf p_s := (w_{s,1}, \cdots, w_{s, n_s})$,  $\mathbf p_t := (w_{t,1}, \cdots, w_{1, n_t})$, and $\mathbf 1_{n_s}$ (respectively $\mathbf 1_{n_t}$) is the unit vector with dim $n_s$ (respectively $n_t$). The discrete transportation plan  $\bm \pi^\star_\theta$ is then defined as:
\begin{equation}
   \bm \pi_\theta^\star := \underset{\bm \pi \in \mathbf \Pi(p_s, p_t)}{\mathrm{argmin}} \langle \bm \pi, \mathbf C_\theta \rangle_F
   \label{equation:solve_OT}
\end{equation}
where $\mathbf C_\theta(i,j) := c_\theta(x_{s,i}, x_{t,j})$ and $\langle \cdot, \cdot \rangle_F$ is the Frobenius dot product. Note that $\bm \pi^\star_\theta$ depends on both $p_s$ and $p_t$, and $\theta$ since $\mathbf C_\theta$ depends on $\theta$. In practice, we use Entropic regularization \parencite{cuturi2013sinkhorn} that makes OT easier to solve by promoting smoother transportation plan with a computationally efficient algorithm, based on Sinkhorn-Knopp’s scaling matrix approach (see the Appendix C).

\subsection{Method}

\begin{algorithm}[t]
\caption{Transported Prototypes. \textcolor{blue}{Blue lines} highlight the OT's contribution in the computational graph of an episode compared to the standard Prototypical Network \parencite{Snell17}.}
\label{alg:transported_prototypes}
 \textbf{Input:} Support set $\mathcal S := (x_{s,i}, y_{s,i})_{1\leq i \leq n_s}$, query set $\mathcal Q := (x_{q,j}, y_{q,j})_{1 \leq j \leq n_q}$, classes $\mathcal C$, backbone $\varphi_\theta$. \\
 \textbf{Output:} Loss $\mathcal L(\theta)$ for a randomly sampled episode. 
\begin{algorithmic}[1] %[1] enables line numbers
\State $z_{s,i}, z_{q,j} \leftarrow \varphi(x_{s,i}; \theta),  \varphi(x_{q,j}; \theta)$, for $i,j$  \Comment{Get representations.}
\textcolor{blue}{
\State $\mathbf C_\theta(i,j) \leftarrow ||z_{s,i} - z_{q,j}||^2$, for $i,j$ \Comment{Cost-matrix.}
\State $\bm \pi_\theta^\star \leftarrow$ Solve Equation \ref{equation:solve_OT} \Comment{Transportation plan.}
\State $\hat{\bm \pi}_\theta^\star(i,j) \leftarrow  \bm \pi^\star_\theta(i,j) / \sum_j \bm \pi_\theta^\star(i,j) $, for $i,j$  \Comment{Normalization.}
\State $\hat{\mathbf S} = (\hat z_{s,i})_i \leftarrow$ Given by Equation \ref{equation:barycenter_mapping} \Comment{Get transported support set.}}
\State $\hat{\mathbf c}_k \leftarrow \frac{1}{|\hat{\mathbf S}_k|} \sum_{\hat z_s \in \hat{\mathbf S}_k} \hat z_s$, for $k \in \mathcal C$.  \Comment{Get \textcolor{blue}{transported} prototypes.}
\State $p_\theta(y| x_{q,j}) \leftarrow $ From Equation \ref{equation:p_theta}, for $j$
\State \textbf{Return:} $\mathcal L(\theta) := \frac{1}{n_q} \sum_{j=1}^{n_q}- \log p_\theta(y_{q,j}| x_{q,j})$.
\end{algorithmic}
\end{algorithm}

\paragraph{Transportation Phase.} At each episode, we are provided with a source support set $\mathcal S$ and a target query set $\mathcal Q$. We note respectively $\mathbf S$ and $\mathbf Q$ their representations from a deep network $\varphi(\cdot; \theta)$ \ie $z_s \in \mathbf S$ is defined as  $z_s := \varphi(x_s;\theta)$ for $x_s \in \mathcal S$, respectively $z_q \in \mathbf Q$ is defined as  $z_q := \varphi(x_q;\theta)$ for $x_q \in \mathcal Q$. As these two sets are sampled from different distributions, $\mathbf S$ and $\mathbf Q$ are likely to lie in different regions of the representation space. In order to adapt the source support set $\mathcal S$ to the target domain, which is only represented by the target query set $\mathcal Q$, we follow \parencite{courty2016optimal} to compute  $\hat{\mathbf S}$ the \textit{barycenter mapping}  of $\mathcal S$, that we refer to as the \textit{transported support set}, defined as follows:
\begin{equation}
    \hat{\mathbf S} := \hat{\bm \pi}_{\theta}^\star \mathbf Q
    \label{equation:barycenter_mapping}
\end{equation}
where $\bm \pi_{\theta}^\star$ is the transportation plan from $\mathbf S$ to $\mathbf Q$ and $\hat {\bm \pi}^\star_\theta := \bm \pi^\star_\theta(i,j) / \sum_{j=1}^{n_t} \bm \pi^\star_\theta(i,j)$. The \textit{transported} support set  $\hat{\mathbf S}$ is an estimation of labelled examples in the target domain using labelled examples in the source domain. The success relies on the fact that transportation conserves labels, \ie a query instance close to $\hat{z_ s} \in  \hat{\mathbf S}$ should share the same label with $x_s$, where $\hat{z_s}$ is the barycenter mapping of $z_s \in \mathbf S$. See  step (3) of Figure \ref{fig:transported_prototypes}  for a visualization of the transportation phase.

\paragraph{Prototype Phase.} For each class $k\in \mathcal C$, we compute the \textit{transported prototypes} $\hat{\mathbf c}_k:= \frac{1}{|\hat{\mathbf S}_k|} \sum_{\hat z_s \in \hat{\mathbf S}_k} \hat z_s$ (where $\hat{\mathbf S}_k$ is the transported support set with class $k$ and $\mathcal C$ are classes of current episode).  We classify each query $x_q$ with representation $z_q= \varphi(x_q; \theta)$ using its euclidean distance to each transported prototypes;
\begin{equation}
    p_\theta(y = k|x_q) := \frac{\exp \left ( - || z_q - \hat{\mathbf{c}}_k ||^2 \right)}{\sum_{k' \in \mathcal C} \exp \left ( - || z_q -  \hat{\mathbf{c}}_{k'} ||^2 \right)} 
    \label{equation:p_theta}
\end{equation}
Crucially, the standard Prototypical Networks \parencite{Snell17} computes euclidean distance to each prototypes while we compute the euclidean to each \textit{transported} prototypes, as presented in step (4) of Figure \ref{fig:transported_prototypes}. Note that our formulation involves the query set in the computation of $(\hat{\mathbf{c}}_k)_{k\in \mathcal C}$.

\paragraph{Genericity of OT.} \texttt{FewShiftBed} implements OT as a stand-alone module that can be easily plugged into any FSL algorithm. We report additional baselines in Appendix B where other FSL algorithms are equipped with OT. This technical choice reflects our insight that OT may be ubiquitous for addressing FSQS and makes its usage in the testbed straightforward.

\section{Experiments}
\label{section:experiments}

\begin{table}[t]
\centering
\resizebox{1.\textwidth}{!}{
\begin{tabular}{l|cc|cc|c|}
\cline{2-6}
                                   & \multicolumn{2}{c|}{\textbf{Meta-CIFAR100-C}} & \multicolumn{2}{c|}{\textbf{miniImageNet-C}} & \textbf{FEMNIST-FS} \\ \cline{2-6} 
                                   & \textbf{1-shot}  & \textbf{5-shot}  & \textbf{1-shot}      & \textbf{5-shot}     & \textbf{1-shot}          \\ \hline
\multicolumn{1}{|l|}{ProtoNet \parencite{Snell17}}     & 30.02 $\pm$ 0.40   & 42.77 $\pm$ 0.47   & 36.37 $\pm$ 0.50                      &         47.58 $\pm$ 0.57           & 84.31 $\pm$ 0.73           \\
\multicolumn{1}{|l|}{MatchingNet \parencite{Vinyals16}}  & 30.71 $\pm$ 0.38   & 41.15 $\pm$ 0.45   &        35.26 $\pm$ 0.50              &     44.75 $\pm$ 0.55                &    84.25 $\pm$ 0.71              \\
\hline 
\multicolumn{1}{|l|}{TransPropNet$\dag$ \parencite{liu2018learning}} & \textbf{34.15 $\pm$ 0.39}   & 47.39 $\pm$ 0.42   &       24.10   $\pm$ 0.27         &                 27.24   $\pm$ 0.33        & 86.42 $\pm$ 0.76           \\
\multicolumn{1}{|l|}{FTNet$\dag$ \parencite{dhillon2019baseline}}        & 28.91 $\pm$ 0.37   & 37.28 $\pm$ 0.40   &        39.02 $\pm$ 0.46     &     51.27 $\pm$ 0.45               & 86.13 $\pm$ 0.71  \\
\multicolumn{1}{|l|}{TP$\dag$ (ours)}   & \textbf{34.00 $\pm$ 0.46}   & \textbf{49.71 $\pm$ 0.47}   &    \textbf{40.49 $\pm$ 0.54}    &       \textbf{59.85 $\pm$ 0.49}                     & \textbf{93.63 $\pm$ 0.63}         \\ \hline \hline
\multicolumn{1}{|l|}{TP w/o OT $\dag$}   & 32.47 $\pm$ 0.41 &  48.00 $\pm$ 0.44 &        40.43 $\pm$ 0.49               &      53.71 $\pm$ 0.50               &  90.36 $\pm$ 0.58  \\
\multicolumn{1}{|l|}{TP w/o TBN $\dag$}  & 33.74 $\pm$ 0.46 &  49.18 $\pm$ 0.49 &    37.32 $\pm$ 0.55                 &        55.16 $\pm$ 0.54             & 92.31 $\pm$ 0.73      \\
\multicolumn{1}{|l|}{TP w. OT-TT $\dag$} & 32.81 $\pm$ 0.46 & 48.62 $\pm$ 0.48 &  \textbf{44.77 $\pm$ 0.57}    &  \textbf{60.46 $\pm$ 0.49}  & \textbf{94.92 $\pm$ 0.55}   \\
\multicolumn{1}{|l|}{TP w/o ET $\dag$}   & \textbf{35.94 $\pm$ 0.45} & 48.66 $\pm$ 0.46 &  42.46    $\pm$ 0.53      &      54.67 $\pm$ 0.48     & 94.22$ \pm$ 0.70  \\ \hline
\multicolumn{1}{|l|}{TP w/o SQS $\dag$}   & \textit{85.67 $\pm$ 0.26} & \textit{88.52 $\pm$ 0.17} &        \textit{64.27 $\pm$ 0.39}    &  \textit{75.22 $\pm$ 0.30}       &   \textit{99.72 $\pm$ 0.07} \\ \hline
\end{tabular}}
\caption{Top-1 accuracy of few-shot learning models in various datasets and numbers of shots with 8 instances per class in the query set (except for FEMNIST-FS: 1 instance per class in the query set), with 95\% confidence intervals. The top half of the table is a comparison between existing few-shot learning methods and Transported Prototypes (TP). The bottom half is an ablation study of TP. OT denotes Optimal Transport, TBN is Transductive Batch-Normalization, OT-TT refers to the setting where Optimal Transport is
applied at test time but not during episodic training, and ET means episodic training
\ie w/o ET refers to the setting where training is performed through standard Empirical Risk Minimization. TP w/o SQS reports model’s performance in the absence of
support-query shift. $\dag$ flags if the method is transductive. For each setting, the best accuracy among existing methods is shown in bold, as well as the accuracy of an ablation if it improves TP.}
\label{table:benchmark}
\end{table}

We compare the performance of baseline algorithms with \textit{Transported Prototypes} on various datasets and settings. We also offer an ablation study in order to isolate the source to the success of \textit{Transported Prototypes}. Extensive results are detailed in Appendix B. Instructions to reproduce these results can be found in the code's documentation.

\paragraph{Setting and details.} We conduct experiments on all methods and datasets implemented in \texttt{FewShiftBed}. We use a standard 4-layer convolutional network for our experiments on Meta-CIFAR100-C and FEMNIST-FewShot, and a ResNet18 for our experiments on miniImageNet. Transductive methods are equipped with a Transductive Batch-Normalization. All episodic training runs contain 40k episodes, after which we retrieve model state with best validation accuracy. We run each individual experiment on three different random seeds. All results presented in this paper are the average accuracies obtained with these random seeds.

\paragraph{Analysis.}  The top half of Table \ref{table:benchmark} reveals that Transported Prototypes (TP) outperform all baselines by a strong margin on all datasets and settings. Importantly, baselines perform poorly on FSQS, demonstrating they are not equipped to address this challenging problem, stressing our study's significance. It is also interesting to note that the performance of transductive approaches, which is significantly better in a standard FSL setting \parencite{liu2018learning,dhillon2019baseline}, is here similar to inductive methods (notably, TransPropNet \parencite{liu2018learning} fails loudly without Transductive Batch-Normalization showing that propagating label with non-overlapping support/query can have a dramatic impact, see Appendix B). Thus, FSQS deserves a fresher look to be solved. Transported Prototypes mitigate a significant part of the performance drop caused by support-query shift while benefiting from the simplicity of combining a popular FSL method with a time-tested UDA method. This gives us strong hopes for future works in this direction.

\paragraph{Ablation study.} Transported Prototypes (TP) combines three components: Optimal Transport (OT), Transductive Batch-Normalization (TBN) and episode training (ET). Which of these components are responsible for the observed gain? Following recommendations from Section \ref{section:protocol}, we ablate those components in the bottom half of Table \ref{table:benchmark}. We observe that both OT and TBN individually improve the performance of ProtoNet for FSQS, and that the best results are obtained when the two of them are combined. Importantly, OT without TBN performs better than TBN without OT (except for 1-shot mIN-C), demonstrating the superiority of OT compared to TBN for aligning distributions in the few samples regime. Note that the use of TaskNorm \parencite{bronskill2020tasknorm} is beyond the scope of the paper\footnote{These normalizations are implemented in \texttt{FewShiftBed} for future works.};  we encourage future work to dig into that direction and we refer the reader to the very recent work \parencite{du2021metanorm}. We observe that there is no clear evidence that using OT at train-time is better than simply applying it at test-time on a ProtoNet trained without OT. Additionally, the value of Episodic Training (ET) compared to standard Empirical Risk Minimization (ERM) is not obvious. For instance, simply training with ERM and applying TP at test-time is better than adding ET on 1-shot MC100-C, 1-shot mIN-C and FEMNIST-FS, making it an another element to add to the study \parencite{laenen2020episodes} who put into question the value of ET. Understanding why and when we should use ET or only OT at test-time is interesting for future works. Additionally, we compare TP with MAP \parencite{hu2020leveraging} which implements an OT-based approach for transductive FSL. Their approach includes a power transform to reduce the skew in the distribution, so for fair comparison we also implemented it into Transported Prototypes for these experiments\footnote{Therefore results in Table \ref{table:TP_MAP} differ from results in Table \ref{table:benchmark}.}. We also used the OT module only at test-time and compared with two backbones, respectively trained with ET and ERM. Interestingly, our experiments in Table \ref{table:TP_MAP} show that MAP is able to handle SQS. Finally, in order to evaluate the performance drop related to Support-Query Shift compared to a setting with support and query instances sampled from the same distribution, we test Transported Prototypes on few-shot classification tasks without SQS (TP w/o SQS in Table \ref{table:benchmark}), making a setup equivalent to CDFSL. Note that in both cases, the model is trained in an episodic fashion on tasks presenting a Support-Query Shift. These results show that SQS presents a significantly harder challenge than CDFSL, while there is considerable room for improvements.

\begin{table}[t]
\resizebox{1.\textwidth}{!}{
\begin{tabular}{l|cc|cc|c|}
\cline{2-6}
        & \multicolumn{2}{c|}{\textbf{Meta-CIFAR100-C}} & \multicolumn{2}{c|}{\textbf{miniImageNet-C}} & \textbf{FEMNIST-FS} \\ 
        \cline{2-6} 
        & \textbf{1-shot}  & \textbf{5-shot}  & \textbf{1-shot}      & \textbf{5-shot}     & \textbf{1-shot}          \\ \hline
\multicolumn{1}{|l|}{TP$^\star$}         & 36.17 $\pm$ 0.47 & 50.45 $\pm$ 0.47 & 45.41 $\pm$ 0.54 & 57.82 $\pm$ 0.48 & 93.60 $\pm$ 0.68 \\
\multicolumn{1}{|l|}{MAP$^\star$}   &  35.96 $\pm$ 0.44 & 49.55 $\pm$ 0.45 & 43.51 $\pm$ 0.47 & 56.10 $\pm$ 0.43 & 92.86 $\pm$ 0.67 \\ \hline 
\multicolumn{1}{|l|}{TP$^\dag$} & 32.13 $\pm$ 0.45 & 46.19 $\pm$ 0.47 & 45.77 $\pm$ 0.58 & 59.91 $\pm$ 0.48 & 94.92 $\pm$ 0.56 \\
\multicolumn{1}{|l|}{MAP$^\dag$} & 32.38 $\pm$ 0.41 & 45.96  $\pm$ 0.43 & 43.81 $\pm$ 0.47 & 57.70 $\pm$ 0.43 & 87.15 $\pm$ 0.66 \\
 \hline
\end{tabular}}
\caption{Top-1 accuracy with 8 instances per class in the query set when applying Transported Prototypes and MAP on two different backbones: $\star$ is standard ERM (\ie without Episodic Training) and $\dag$ is ProtoNet \parencite{Snell17}. Transported Prototypes performs equally or better than MAP \parencite{hu2020leveraging}. Here TP includes power transform in the feature space.}
\label{table:TP_MAP}
\end{table}

\section{Conclusion}
\label{section:conclusion}
We release \texttt{FewShiftBed}, a testbed for the under-investigated and crucial problem of Few-Shot Learning when the support and query sets are sampled from related but different distributions, named FSQS. \texttt{FewShiftBed} includes three datasets, relevant baselines and a protocol for reproducible research. Inspired from recent progress of Optimal Transport (OT) to address Unsupervised Domain Adaptation, we propose a method that efficiently combines OT with the celebrated Prototypical Network \parencite{Snell17}. Following the protocol of \texttt{FewShiftBed}, we bring compelling experiments demonstrating the advantage of our proposal compared to transductive counterparts. We also isolate factors responsible for improvements. Our findings suggest that Batch-Normalization is ubiquitous, as described in related works \parencite{bronskill2020tasknorm,du2021metanorm}, while episodic training, even if promising on paper, is questionable. As a lead for future works, \texttt{FewShiftBed} could be improved by using different datasets to model different domains, instead of using artificial transformations. Since we are talking about domain adaptation, we also encourage the study of accuracy as a function of the size of the target domain, \ie the size of the query set. Moving beyond the transductive algorithm, as well as understanding when meta-learning brings a clear advantage to address FSQS remains an open and exciting problem. \texttt{FewShiftBed} brings the first step towards its progress.

\section*{Acknowledgements}
Etienne Bennequin is funded by Sicara and ANRT (France), and Victor Bouvier is funded by Sidetrade and ANRT (France), both through a CIFRE collaboration with CentraleSup\'elec. This work was performed using HPC resources from the “M\'esocentre” computing center of CentraleSup\'elec and \'Ecole Normale Sup\'erieure Paris-Saclay supported by CNRS and R\'egion \^Ile-de-France (\url{http://mesocentre.centralesupelec.fr/}).

\printbibliography

\begin{refsection}

\appendix
 \section{Extended positioning}

\paragraph{Few-Shot Classification.} Methods to solve the Few-Shot Classification problem \parencite{lake2011} are usually put into one of these three categories \parencite{Chen19}: metric-based, optimization-based, and hallucination-based. Most metric-learning methods are built on the principle of Siamese Networks \parencite{Koch15}, while also exploiting the meta-learning paradigm: they learn a feature extractor across training tasks \parencite{Vinyals16}. Prototypical Networks \parencite{Snell17} classify queries from their euclidean distances to one prototype embedding per class. Relation Networks \parencite{Sung18} add an other deep network on top of Prototype Networks to replace the euclidean distance. Optimization-based methods use an other approach: learning to fine-tune. MAML \parencite{Finn17} and Reptile \parencite{nichol2018firstorder} learn a good model initialization, \textit{i.e.} model parameters that can adapt to a new task (with novel classes) in a small number of gradient steps. Other methods such as Meta-LSTM \parencite{Ravi16} and Meta-Networks \parencite{Munkhdalai17} replace standard gradient descent by a meta-learned optimizer. Hallucination-based methods aim at augmenting the scarce labeled data, by hallucinating feature vectors \parencite{hariharan2017low}, using Generative Adversarial Networks \parencite{antoniou2017data}, or meta-learning \parencite{wang2018low}. Recent works also suggest that competitive results in Few-Shot Classification can be achieved with more simple methods based on fine-tuning \parencite{Chen19,goldblum2020unraveling}.

\paragraph{Transductive Few-Shot Classification.} Some methods aim at solving few-shot classification tasks by using the query set as unlabeled data. Transductive Propagation Network \parencite{liu2018learning} meta-learns label propagation from the support to query set concurrently with the feature extractor. Antoniou \& Storkey \parencite{antoniou2019learning} proposed to use a meta-learned critic network to further adapt a classifier on the query set in an unsupervised setting. Ren \textit{et al.} \parencite{ren2018meta} extend Prototypical Networks in order to use the query set in the prototype computation. Transductive Information Maximization \parencite{boudiaf2020transductive} aims at maximizing the mutual information between the features extracted from the query set and their predicted labels. Finally, Transductive Fine-Tuning \parencite{dhillon2019baseline} augments standard fine-tuning using the classification entropy of all query instances.

\paragraph{Unsupervised Domain Adaptation.} UDA has a long standing story \parencite{pan2009survey,quionero2009dataset}. The analysis of the role of representations from \parencite{ben2007analysis} has led to wide literature based on domain invariant representations \parencite{ganin2015unsupervised,long2015learning}. Outstanding progress have been towards learning more domain transferable representations by looking for domain invariance. The tensorial product between representations and prediction promotes conditional domain invariance \parencite{long2018conditional}, the use of weights \parencite{cao2018partial,you2019universal,bouvier2020robust,combes2020domain} has dramatically improved the problem of label shift theoretically described in \parencite{zhang2019bridging}, hallucinating consistent target samples \parencite{liu2019transferable}, penalizing high singular values of batch of representations \parencite{chen2019transferability} or by enforcing the favorable inductive bias of consistence through various data augmentation in the target domain \parencite{ouali2020target}. Recent works address the problem of adaptation without source data \parencite{liang2020we,yehsofa}. The seminal work \parencite{courty2016optimal}, followed by  \parencite{courty2017joint,bhushan2018deepjdot}, brings Optimal Transport (OT) to UDA by transporting source samples in the target domain.

\paragraph{Test-Time Adaptation.} Test-time Adaptation (TTA) is the subject of recent pioneering works. In \parencite{sun2020test}, adaptation is performed by test-time training of representations through a self-supervision task which consists in predicting the rotation of an image. This leads to a successful adaptation when the gradient of fine-tuning procedure is correlated with the gradient of the cross-entropy between the prediction and the label of the target sample, which is not available. Inspired from UDA methods based on domain invariance of representations, a line of works \parencite{nado2020evaluating,schneider2020improving} aims to align the mean and the variance of train and test distribution of representations. This is simply done by updating statistics of the batch-normalization layer. In a similar spirit of leveraging the batch-normalization layer for adaptation, \parencite{wang2020fully} suggests to minimize prediction entropy on a batch of test samples, as suggested in semi-supervised learning \parencite{grandvalet2005semi}. As pointed by authors of \parencite{wang2020fully}, updating the whole network is inefficient and exposes to a risk of test batch overfit. To adress this problem, authors suggest to only update batch-normalization parameters for minimizing prediction's entropy. The paradigm of \textit{Adaptative Risk Minimization} (ARM) is introduced in \parencite{zhang2020adaptive}. ARM aims to adapt a classifier at test-time by conditioning its prediction on the whole batch of test samples (not only one sample). Authors demonstrate that such classifier is meta-trainable as long as the training data exposes a structure of group. Consequently, \parencite{zhang2020adaptive} is closer work to ours, while we have more ambitious perspectives as we address the problem of few-shot learning \ie few-shot are available per class while new classes are discovered at test-time.

\paragraph{Few-Shot Classification under Distributional Shift.} Recent works on few-shot classification tackle the problem of distributional shift between the meta-training set and the meta-testing set. Chen \textit{et al.} \parencite{Chen19} compare the performance of state-of-the-art solutions to few-shot classification on a cross-domain setting (meta-training on \textit{mini}ImageNet \parencite{Vinyals16} and meta-testing on Caltech-UCSD Birds 200 \parencite{WelinderEtal2010}). Zhao \textit{et al.} propose a Domain-Adversarial Prototypical Network \parencite{zhao2020domain} in order to both align source and target domains in the feature space and maintain discriminativeness between classes. Considering the problem as a shift in the distribution of tasks (\textit{i.e.} training and testing tasks are drawn from two distinct distributions), Sahoo \textit{et al.} combine Prototypical Networks with adversarial domain adaptation at task level \parencite{sahoo2019meta}. While these works address the key issue of distributional shift between meta-training and meta-testing, they assume that for each task, the support set and query set are always drawn from the same distribution. We find that this assumption rarely holds in practice. In this work we consider a distributional shift both between meta-training and meta-testing and between support and query set.

\section{All experimental results}
\label{section:appendix-results}

In this section we present the extended results of our experiments. Prototypical Networks, Matching Networks and Transductive Propagation Networks have been declined in 10 distinct versions:
\begin{itemize}
    \item Original algorithms: \textbf{episodic training}, with Conventional Batch-Normalization (\textbf{CBN}) and not Optimal Transport (\textbf{Vanilla});
    \item \textbf{Episodic training} and \textbf{CBN}, with Optimal Transport applied at test time (\textbf{OT-TT});
    \item \textbf{Episodic training }and \textbf{CBN}, with Optimal Transport integrated into the algorithm both during training and testing (\textbf{OT});
    \item \textbf{Episodic training}, with Transductive Batch-Normalization (\textbf{TBN}) and not Optimal Transport (\textbf{Vanilla});
    \item \textbf{Episodic training} and \textbf{TBN}, with \textbf{OT-TT};
    \item \textbf{Episodic training} and \textbf{TBN}, with \textbf{OT};
    \item Standard Empirical Risk Minimization (\textbf{ERM}) instead of episodic training, with \textbf{CBN} and not Optimal Transport (\textbf{Vanilla});
    \item \textbf{ERM} with \textbf{CBN} and \textbf{OT};
    \item \textbf{ERM} with \textbf{TBN} and no Optimal Transport (\textbf{Vanilla});
    \item \textbf{ERM} with \textbf{TBN} and \textbf{OT}.
\end{itemize}

Transductive Fine-Tuning (FTNet) is not compatible with episodic training. Also the integration of Optimal Transport into this algorithm is non trivial. Therefore we only applied FTNet with ERM and without OT.

Every result presented in the following tables is the average over three runs with three random seeds (\texttt{1}, \texttt{2} and \texttt{3}). For clarity, we do not report the 95\% confidence interval for each result. Keep in mind that this interval is different for each result, but we found that it is always greater than $\pm$ 0.2\% and smaller than $\pm$ 0.8\%.

Details of the experiments and instructions to reproduce them are available in the code.
\begin{table}
\begin{tabular}{l|ccc|ccc|cc|cc|}
\cline{2-11}
\multirow{4}{*}{}                 & \multicolumn{10}{c|}{\textbf{Meta-CIFAR100-C 1-shot 8-target}}                                                                                                                                                            \\ \cline{2-11} 
                                  & \multicolumn{6}{c|}{Episodic training}                                                                                                      & \multicolumn{4}{c|}{Standard ERM}                                           \\ \cline{2-11} 
                                  & \multicolumn{3}{c|}{CBN}                                             & \multicolumn{3}{c|}{TBN}                                             & \multicolumn{2}{c|}{CBN}             & \multicolumn{2}{c|}{TBN}             \\ \cline{2-11} 
                                  & \multicolumn{1}{c}{Vanilla} & \multicolumn{1}{c}{w. OT-TT} & w. OT & \multicolumn{1}{c}{Vanilla} & \multicolumn{1}{c}{w. OT-TT} & w. OT & \multicolumn{1}{c}{Vanilla} & w. OT & \multicolumn{1}{c}{Vanilla} & w. OT \\ \hline
\multicolumn{1}{|l|}{ProtoNet}    & 30.02                        & 32.11                         & 33.74 & 32.47                        & 32.81                         & 34.00 & 29.10                        & 35.48 & 29.79                        & 35.4  \\
\multicolumn{1}{|l|}{MatchingNet} & 30.71                        & 32.85                         & 34.48 & 32.97                        & 32.78                         & 35.11 & 33.50                        & \textbf{36.13} & 33.67                        & \textbf{35.87} \\
\multicolumn{1}{|l|}{PropNet}     & 30.26                        & 28.70                          & 26.87 & 34.15                        & 29.48                         & 27.68 & 23.33                        & 31.08 & 22.55                        & 31.20 \\
\multicolumn{1}{|l|}{FTNet}       &                              &                               &       &                              &                               &       & 28.91                        &       & 28.75                        &       \\ \hline
\end{tabular}
\caption{Ablation for Meta-CIFAR100-C 1-shot 8-target.}
\end{table}

\begin{table}
\begin{tabular}{l|ccc|ccc|cc|cc|}
\cline{2-11}
\multirow{4}{*}{}                 & \multicolumn{10}{c|}{\textbf{Meta-CIFAR100-C 1-shot 16-target}}                                                                                                                                                           \\ \cline{2-11} 
                                  & \multicolumn{6}{c|}{Episodic training}                                                                                                      & \multicolumn{4}{c|}{Standard ERM}                                           \\ \cline{2-11} 
                                  & \multicolumn{3}{c|}{CBN}                                             & \multicolumn{3}{c|}{TBN}                                             & \multicolumn{2}{c|}{CBN}             & \multicolumn{2}{c|}{TBN}             \\ \cline{2-11} 
                                  & \multicolumn{1}{c}{Vanilla} & \multicolumn{1}{c}{w. OT-TT} & w. OT & \multicolumn{1}{c}{Vanilla} & \multicolumn{1}{c}{w. OT-TT} & w. OT & \multicolumn{1}{c}{Vanilla} & w. OT & \multicolumn{1}{c}{Vanilla} & w. OT \\ \hline
\multicolumn{1}{|l|}{ProtoNet}    & 29.98                        & 32.24                         & 35.63 & 32.52                        & 31.72                         & \textbf{36.20}  & 29.02                        & 35.89 & 29.61                        & 35.94 \\
\multicolumn{1}{|l|}{MatchingNet} & 31.1                         & 30.94                         & 35.53 & 33.08                        & 33.28                         & \textbf{36.36} & 33.49                        & \textbf{36.61} & 33.64                        & \textbf{36.54} \\
\multicolumn{1}{|l|}{PropNet}     & 30.82                        & 32.39                         & 31.15 & 34.83                        & 33.53                         & 31.33 & 26.81                        & 33.9  & 27.92                        & 34.10  \\
\multicolumn{1}{|l|}{FTNet}       &                              &                               &       &                              &                               &       & 29.01                        &       & 28.86                        &       \\ \hline
\end{tabular}
\caption{Ablation for Meta-CIFAR100-C 1-shot 16-target.}
\end{table}

\begin{table}
\begin{tabular}{l|ccc|ccc|cc|cc|}
\cline{2-11}
\multirow{4}{*}{}                 & \multicolumn{10}{c|}{\textbf{Meta-CIFAR100-C 5-shot 8-target}}                                                                                                                                                            \\ \cline{2-11} 
                                  & \multicolumn{6}{c|}{Episodic training}                                                                                                      & \multicolumn{4}{c|}{Standard ERM}                                           \\ \cline{2-11} 
                                  & \multicolumn{3}{c|}{CBN}                                             & \multicolumn{3}{c|}{TBN}                                             & \multicolumn{2}{c|}{CBN}             & \multicolumn{2}{c|}{TBN}             \\ \cline{2-11} 
                                  & \multicolumn{1}{c}{Vanilla} & \multicolumn{1}{c}{w. OT-TT} & w. OT & \multicolumn{1}{c}{Vanilla} & \multicolumn{1}{c}{w. OT-TT} & w. OT & \multicolumn{1}{c}{Vanilla} & w. OT & \multicolumn{1}{c}{Vanilla} & w. OT \\ \hline
\multicolumn{1}{|l|}{ProtoNet}    & 42.77                        & 47.54                         & 48.37 & 48.00                        & 48.62                         & \textbf{49.71} & 44.89                        & 48.61 & 46.59                        & 48.66 \\
\multicolumn{1}{|l|}{MatchingNet} & 41.15                        & 43.90                          & 44.55 & 45.05                        & 44.86                         & 45.78 & 43.00                           & 45.35 & 43.51                        & 45.10  \\
\multicolumn{1}{|l|}{PropNet}     & 39.13                        & 40.60                          & 25.68 & 47.39                        & 40.47                         & 27.29 & 29.32                        & 39.82 & 29.50                         & 29.82 \\
\multicolumn{1}{|l|}{FTNet}       &                              &                               &       &                              &                               &       & 37.28                        &       & 37.40                         &       \\ \hline
\end{tabular}
\caption{Ablation of Meta-CIFAR100-C 5-shot 8-target.}
\end{table}

\begin{table}
    \begin{tabular}{l|ccc|ccc|cc|cc|}
    \cline{2-11}
    \multirow{4}{*}{}                 & \multicolumn{10}{c|}{\textbf{Meta-CIFAR100-C 5-shot 16-target}}                                                                                                                                                           \\ \cline{2-11} 
                                      & \multicolumn{6}{c|}{Episodic training}                                                                                                      & \multicolumn{4}{c|}{Standard ERM}                                           \\ \cline{2-11} 
                                      & \multicolumn{3}{c|}{CBN}                                             & \multicolumn{3}{c|}{TBN}                                             & \multicolumn{2}{c|}{CBN}             & \multicolumn{2}{c|}{TBN}             \\ \cline{2-11} 
                                      & \multicolumn{1}{c}{Vanilla} & \multicolumn{1}{c}{w. OT-TT} & w. OT & \multicolumn{1}{c}{Vanilla} & \multicolumn{1}{c}{w. OT-TT} & w. OT & \multicolumn{1}{c}{Vanilla} & w. OT & \multicolumn{1}{c}{Vanilla} & w. OT \\ \hline
    \multicolumn{1}{|l|}{ProtoNet}    & 42.07                        & 48.26                         & 48.25 & 46.49                        & 48.71                         & \textbf{49.94} & 44.67                        & 48.61 & 46.48                        & 48.89 \\
    \multicolumn{1}{|l|}{MatchingNet} & 41.74                        & 44.51                         & 45.71 & 44.91                        & 44.71                         & 47.37 & 42.97                        & 46.06 & 46.22                        & 46.37 \\
    \multicolumn{1}{|l|}{PropNet}     & 38.73                        & 39.25                         & 37.22 & 43.91                        & 40.62                         & 40.02 & 33.06                        & 40.03 & 33.93                        & 40.03 \\
    \multicolumn{1}{|l|}{FTNet}       &                              &                               &       &                              &                               &       & 37.51                        &       & 37.66                        &       \\ \hline
    \end{tabular}
    \caption{Ablation of Meta-CIFAR100-C 5-shot 16-target.}
\end{table}

\begin{table}
    \begin{tabular}{l|ccc|ccc|cc|cc|}
    \cline{2-11}
    \multirow{4}{*}{}                 & \multicolumn{10}{c|}{\textbf{FEMNIST-FewShot 1-shot 1-target}}                                                                                                                                                            \\ \cline{2-11} 
                                      & \multicolumn{6}{c|}{Episodic training}                                                                                                      & \multicolumn{4}{c|}{Standard ERM}                                           \\ \cline{2-11} 
                                      & \multicolumn{3}{c|}{CBN}                                             & \multicolumn{3}{c|}{TBN}                                             & \multicolumn{2}{c|}{CBN}             & \multicolumn{2}{c|}{TBN}             \\ \cline{2-11} 
                                      & \multicolumn{1}{c|}{Vanilla} & \multicolumn{1}{c|}{w. OT-TT} & w. OT & \multicolumn{1}{c|}{Vanilla} & \multicolumn{1}{c|}{w. OT-TT} & w. OT & \multicolumn{1}{c|}{Vanilla} & w. OT & \multicolumn{1}{c|}{Vanilla} & w. OT \\ \hline
    \multicolumn{1}{|l|}{ProtoNet}    & 84.31                        & 94.00                            & 92.31 & 90.36                        & \textbf{94.92}                         & 93.63 & 80.20                         & 94.30  & 86.22                        & 94.22 \\
    \multicolumn{1}{|l|}{MatchingNet} & 84.25  & 93.66  & 92.73 & 91.05 & \textbf{95.37} & 93.62 & 85.04                        & 94.34 & 87.19                        & 94.26 \\
    \multicolumn{1}{|l|}{PropNet}     & 31.30                         & 40.60                          & 79.30  & 86.42                        & 93.08                         & 87.52 & 45.36                        & 73.64 & 47.34                        & 79.50  \\
    \multicolumn{1}{|l|}{FTNet}       &                              &                               &       &                              &                               &       & 86.13                        &       & 85.92                        &       \\ \hline
    \end{tabular}
    \caption{Ablation of FEMNIST-FewShot 1-shot 1-target.}
\end{table}

\begin{table}[ht]
\centering
\resizebox{.7\textwidth}{!}{

\begin{tabular}{l|cc|cc|c|}
\cline{2-6}
& \multicolumn{2}{c|}{\textbf{Meta-CIFAR100-C}} & \multicolumn{2}{c|}{\textbf{miniImageNet-C}} & \textbf{FEMNIST-FS} \\ \cline{2-6} 
    & \textbf{1-shot}  & \textbf{5-shot}  & \textbf{1-shot}      & \textbf{5-shot}     & \textbf{1-shot}          \\ \hline
\multicolumn{1}{|l|}{MAP}         & 36.58    & 49.37    &  43.38    &  56.25           & 92.94 \\
\multicolumn{1}{|l|}{TP (ours)}   & 36.51 &  50.60  &   45.38   &      61.46               &  93.63  \\
\hline
\end{tabular}}
\caption{Top-1 accuracy of MAP \parencite{hu2020leveraging} compared to Transported Prototypes (ours). Both methods incorporate Optimal Transport into Few-Shot Learning. MAP \parencite{hu2020leveraging} is originally designed for standard transductive FSL. Interestingly, MAP and TP perform quite similarly demonstrating that OT is a powerful tool for addressing FSQS. Note that MAP leverages a Power-Transform that we also plug in TP for comparison, resulting in a boost of performance. Understanding which learners operate best with Optimal Transport is an exciting question. In particular, by proposing TP, we have shown that we result in a strong, interpretable and theoretically motivated method by following principles when applying OT in UDA.} 
\end{table}

\section{Training details}

Entropic regularization for Optimal Transport was proposed in \parencite{cuturi2013sinkhorn} and makes OT easier to solve. It is defined as $
    \bm \pi^\star_\theta(\hat p_s,\hat p_t) := \arg \min_{\bm \pi \in \mathbf \Pi} \langle \bm \pi, \mathbf C_\theta \rangle_F + \varepsilon \Omega(\bm \pi)
$ with $\varepsilon>0$ and $\Omega(\bm \pi) = \sum_{i,j=1}^{n_s, n_t} \bm \pi(i,j)\log \bm \pi(i,j)$ is the negative entropy. It promotes smoother transportation plan while allowing to derive a computationally efficient algorithm, based on Sinkhorn-Knopp’s scaling matrix approach \parencite{knight2008sinkhorn}. In our experiment, we set $\varepsilon=0.05$, but it is possible to tune it, eventually meta-learning it.

\printbibliography

\end{refsection}

\end{document}